\documentclass{article}

\usepackage{graphicx}

\usepackage{nac_preprint}
\usepackage[utf8]{inputenc} 
\usepackage[T1]{fontenc}    
\usepackage{hyperref}       
\usepackage{url}            
\usepackage{booktabs}       
\usepackage{amsfonts}       
\usepackage{nicefrac}       
\usepackage{microtype}      

\usepackage{booktabs} 
\usepackage{subcaption}
\usepackage{amsmath}
\usepackage{amssymb}
\usepackage{cleveref}
\usepackage{mathtools}
\usepackage[toc,page]{appendix}
\usepackage{enumitem}
\usepackage{graphics}
\usepackage{graphicx}
\usepackage{xcolor}
\usepackage[export]{adjustbox}
\usepackage{algorithm}
\usepackage{algorithmicx}
\usepackage{multicol}
\usepackage[noend]{algpseudocode}

\usepackage{textcomp}
\usepackage{diagbox} 
\usepackage{caption}
\usepackage{color, colortbl} 
\definecolor{LightCyan}{rgb}{0.88,1,1}
\definecolor{LightOrange}{rgb}{1,0.76,0.86}

\algnewcommand{\LineComment}[1]{\State \(//\) #1}
\algnewcommand{\RLineComment}[1]{\State \(\triangleright\) #1}
\newlength{\algrhswidth}
\usepackage{bm}
\usepackage{multirow}

\usepackage{wrapfig}

\usepackage{etoolbox}
\usepackage{tikz}
\usetikzlibrary{tikzmark}
\usetikzlibrary{calc}

\errorcontextlines\maxdimen

\newcommand{\ALGtikzmarkcolor}{black}
\newcommand{\ALGtikzmarkextraindent}{4pt}
\newcommand{\ALGtikzmarkverticaloffsetstart}{-.5ex}
\newcommand{\ALGtikzmarkverticaloffsetend}{-.5ex}
\makeatletter
\newcounter{ALG@tikzmark@tempcnta} 

\newcommand\ALG@tikzmark@start{%
    \global\let\ALG@tikzmark@last\ALG@tikzmark@starttext%
    \expandafter\edef\csname ALG@tikzmark@\theALG@nested\endcsname{\theALG@tikzmark@tempcnta}%
    \tikzmark{ALG@tikzmark@start@\csname ALG@tikzmark@\theALG@nested\endcsname}%
    \addtocounter{ALG@tikzmark@tempcnta}{1}%
}

\def\ALG@tikzmark@starttext{start}
\newcommand\ALG@tikzmark@end{%
    \ifx\ALG@tikzmark@last\ALG@tikzmark@starttext
    \else
        \tikzmark{ALG@tikzmark@end@\csname ALG@tikzmark@\theALG@nested\endcsname}%
        \tikz[overlay,remember picture] \draw[\ALGtikzmarkcolor] let \p{S}=($(pic cs:ALG@tikzmark@start@\csname ALG@tikzmark@\theALG@nested\endcsname)+(\ALGtikzmarkextraindent,\ALGtikzmarkverticaloffsetstart)$), \p{E}=($(pic cs:ALG@tikzmark@end@\csname ALG@tikzmark@\theALG@nested\endcsname)+(\ALGtikzmarkextraindent,\ALGtikzmarkverticaloffsetend)$) in (\x{S},\y{S})--(\x{S},\y{E});%
    \fi
    \gdef\ALG@tikzmark@last{end}%
}

\apptocmd{\ALG@beginblock}{\ALG@tikzmark@start}{}{\errmessage{failed to patch}}
\pretocmd{\ALG@endblock}{\ALG@tikzmark@end}{}{\errmessage{failed to patch}}
\makeatother

\title{Deep Domain Adaptation: A Sim2Real Neural Approach for Improving Eye-Tracking Systems}

\author{%
Viet Dung Nguyen\\
 Rochester Institute of Technology \\ 
\texttt{vn1747@rit.edu}
\And 
Reynold Bailey\\
 Rochester Institute of Technology \\ 
\texttt{rjbvcs@rit.edu}
\And 
Gabriel J. Diaz\\
 Rochester Institute of Technology \\ 
\texttt{Gabriel.Diaz@rit.edu}
\And 
Chengyi Ma\\
 Rochester Institute of Technology \\ 
\texttt{cxm3593@rit.edu}
\And 
Alexander Fix\\
 Meta Reality Labs Research \\ 
\texttt{alexander.fix@meta.com}
\And 
Alexander Ororbia \\
Rochester Institute of Technology \\
\texttt{ago@cs.rit.edu}
}

\begin{document}

\setlength{\abovedisplayskip}{0.065cm}
\setlength{\belowdisplayskip}{0pt}

\maketitle

\begin{abstract}
Eye image segmentation is a critical step in eye tracking that has great influence over the final gaze estimate. Segmentation models trained using supervised machine learning can excel at this task, their effectiveness is determined by the degree of overlap between the narrow distributions of image properties defined by the target dataset and highly specific training datasets, of which there are few. Attempts to broaden the distribution of existing eye image datasets through the inclusion of synthetic eye images have found that a model trained on synthetic images will often fail to generalize back to real-world eye images. In remedy, we use dimensionality-reduction techniques to measure the overlap between the target eye images and synthetic training data, and to prune the training dataset in a manner that maximizes distribution overlap. We demonstrate that our methods result in robust, improved performance when tackling the discrepancy between simulation and real-world data samples.
\end{abstract}

\section{Introduction}
\label{sec:intro}

Research in semantic segmentation has a wide range of applications, including autonomous vehicles, medical image analysis, and virtual reality~\cite{Minaee2020ImageSegmentationSurvey}. In the context of eye-tracking, the ability to accurately segment the eye's features provides great utility for the task of gaze estimation. Most modern approaches to eye-tracking rely on different segmented features of the eye, including the iris or pupil centroid or boundary~\cite{Ghosh2021ETSurvey}. For instance, some schemes for estimating and tracking gaze dynamics requires access to iris features uncovered through the use of its texture and velocity~\cite{gaze-iris-1} while others do so by extracting them according to color thresholding and ellipse fitting~\cite{iris_1}, multi-grid search via gradient ascent, and 2-D Gabor filters~\cite{iris_2}. In addition, segmenting different parts of the eye simultaneously enables us to perform center localization, elliptical contour estimation, and blink detection~\cite{DeepVOG}. Ultimately, eye segmentation serves as an essential component of the general modeling toolbox for the eye-tracking community at large.

In general, approaches to eye segmentation~\cite{ritnet, ellseg, ellseg-gen, eye-seg-cnn-1, eye-seg-cnn-2, eye-seg-cnn-3} leverage deep convolutional neural networks~\cite{deep-learning-book, lecun1995convolutional, conv} (CNNs) and, consequently, require large eye datasets in order to train these neural models effectively. The requisite datasets can be collected by recording synthetic information from simulations~\cite{riteyes} or from human subjects directly~\cite{openeds, mpii}. Although, real-world eye datasets, such as OpenEDS~\cite{openeds} or MPIIGaze~\cite{mpii}, provide invaluable samples of data/images to train the CNN models on, constructing such datasets requires a great deal of human annotation effort (high labeling burden) as well as introduces potential human subject image privacy issues. In contrast, synthetic eye datasets circumvent these issues, reducing the data collection effort inherent to working with actual human participants as well as manual labeling work needed to generate ground truth segmentation masks~\cite{riteyes}. As a result, generating datasets of synthetic data samples offers the potential to train powerful eye-tracking CNN-based systems at greatly reduced overall cost.

Despite the promise that synthetic data brings with it, a major issue emerges -- due to the imperfections underlying computer simulation and 3D graphics models, a ``reality gap'' or mismatch exists between the synthetic data produced by a simulated environment (i.e. synthetic eye images) and the real world~\cite{sim2real, sim2real2} (i.e. images of real eyes). Fundamentally, this mismatch (known as the \textit{sim2real} problem) is caused by a domain gap or domain distribution shift which results from a violation of the \textit{independent and identically distributed (iid)} assumption that drives much of modern-day machine learning; the simulator represents the training distribution of the eye-tracking CNN system and the real-world represents a test distribution that the simulator can only at best approximate. 
This distributional mismatch results in degraded test-time generalization ability when the neural model is trained using data from the simulator but evaluated in the real world. For example, past work has suggested that training eye-tracking CNN models with multiple eye datasets potentially degrades their segmentation performance for a particular within-domain dataset~\cite{ellseg-gen}. In this work, we will address this imposing \textit{sim2real} problem in order to improve the accuracy of the segmentation ability of neural eye-tracking models. 
Specifically, our research will improve the performance of a segmentation neural network on an eye dataset, consisting of actual real-world image patterns, training it using synthetic eye datasets and only a small number of real images. In this work, we will utilize the OpenEDS eye segmentation dataset~\cite{openeds} as the real-world dataset, which contains segmentation maps for each part of the eye (manually labeled), and RITEyes as the synthetic dataset, where the label/segmentation map is programmatically generated~\cite{riteyes}. 

The key contributions of this work are as follows: 
\textbf{1)} in closing the domain gap between synthetically generated eye images and real eye images, our approach will ensure that the eye-tracking CNN model is trained with a large number of synthetically generated images in proportion to real-world images by leveraging a learned neural distance model, resulting in little to no degradation of segmentation performance on the real-world test dataset, and 
\textbf{2)} empirically, we will demonstrate that our scheme results in overall higher generalization performance, with respect to mean intersection over union (Jaccard Index) \cite{jaccard}, compared to baseline models trained on synthetic images only. 

\section{Related Work}
\label{sec:lit_review}

\subsection{Segmentation through Deep Learning}
\label{sec:segnets}

A critical component of an image segmentation system is the segmentation network. Concretely, a segmentation network is either a parameterized multi-layer perceptron (MLP)~\cite{Haykin1994NeuralNetworks} or convolution neural network (CNN)~\cite{deep-learning-book, lecun1995convolutional}. In our context, the segmentation network takes in an eye image and produces a classification of each pixel (as `pupil', `iris', `sclera', or `background/other'); this is often referred to as the segmentation map~\cite{unet}. Note that a segmentation network's input and output have the same width and height dimension and, thus, it is designed to be an encoder-decoder architecture and, normally, the exact structure is based on the U-net form~\cite{unet}. Research in eye tracking further extends the segmentation network form/design to better fit within the eye tracking context, introducing additional task-specific objective functions, e.g., as in RITnet~\cite{ritnet} or Ellseg~\cite{ellseg}.

\textbf{U-net Architecture.} As mentioned above, the U-net architecture is a popular method for generating an image segmentation \cite{unet}. It is designed to predict the probability of multiple segmentation classes that each pixel within the image could fall under. The U-net encoder-decoder neural architecture specifically designs the encoder such that each layer has a synaptic skip connection to the corresponding layer in the decoder. This skip connection involves the concatenation of the encoder layer output with the upsampled feature from its same-level decoder layer. The synaptic skip connections provide contextual information with respect to the current layer's image resolution, allowing them to consider the context from the macro to micro-features of the image itself \cite{unet}. This makes the U-net architecture quite effective in, and appropriate for tasks related to image segmentation.

\textbf{RITnet.} Specific to the context of eye-tracking, Chaudhary et al. developed an efficient real-time eye segmentation model known as RITnet \cite{ritnet}, which utilizes a U-net architecture in tandem with several additional objective/loss functions that focus on segmenting particular eye regions. The first loss function in RITnet is the generalized dice loss (GDL), which combines the weighted dice score (this measures the overlap of the prediction and ground truth coefficient) with the cross-entropy loss; this pairing results in improved stability with respect to the cross-entropy objective. The second loss is the boundary aware loss (BAL), which weights the loss in each pixel by its distance to the nearest edge. To achieve this, a mask is computed by dilating the edge using the Canny edge detector\cite{canny1986computational}; the mask is then used to weight the original cross-entropy loss, thus maximizing the correct pixel near each boundary. Finally, the RITnet framework introduces the surface loss, which scales the loss value at each pixel with respect to the pixel's distance to the boundary of the corresponding segmentation class. Specifically, for each segmentation class, a heat map of distance is computed by assigning to each pixel the relative distance to the nearest boundary of the corresponding class. The final surface loss is achieved by averaging the product of the neural network segmentation output and the surface heat map for each segmentation class.

\subsection{Generative Adversarial Learning in Domain Transfer}
\label{sec:gan_transfer}

This work's approach to improving segmentation performance will rely on synthetic data and, in order to ensure the synthetic data is useful, we will craft a scheme that will make the (in-)distribution of the training data closer to the real (out-)data distribution of real eye images. 
More specifically, our approach could be likened to a refinement process that takes in output from an eye simulator (e.g. synthetic images produced by Blender) and modifies it to produce images that are closer to the distribution of the real images. One possible way that we could implement this refinement is through a generative method, such as histogram matching~\cite{histogram-matching-1, histogram-matching-2} where the pixel values in the source image are adjusted so that the histogram of the source and target images match one another. However, changing the features directly, such as pixel values, to match the raw image statistics, e.g., feature histograms, results in a difficult and complex image distributional modeling problem. Additionally, histogram matching can potentially introduce noise to the output image, increase the contrast, or distort the structure of the images~\cite{histogram-matching-problem}. In contrast to generative methodology based on statistic matching, generative approaches based on unsupervised deep neural networks, specifically the generative adversarial network (GAN)~\cite{gan} and variants such as the CycleGAN~\cite{cycle-gan}, offer a parameterized means of mapping from source in-distribution to a target dataset (out-)distribution. Given the modeling flexibility afforded by neural models, our refinement process will leverage and build on the GAN as a central component.

\textbf{Generative Adversarial Network (GAN).} 
A GAN essentially consists of a ``generator'' and a ``discriminator'' neural network that work together to perform unsupervised image generation. The generator takes in as input noise, i.e., a noisy latent vector, and outputs a synthesized image pattern that looks similar to those in the desired image space. The discriminator specifically tries to classify whether an image is the output of the generator (fake) or the real image (real). The objective of the generator is to generate images that are plausible enough so as to reduce the accuracy of (or ``fool'') the discriminator. The objective of the discriminator $D$ is to maximize the binary cross-entropy loss for data coming from both the real image domain and the generated image domain.

\textbf{CycleGAN.} The CycleGAN \cite{cycle-gan}, which is an extension of the basic GAN, is built with the goal of facilitating image translation across different (input) data domains. Given an image from one data domain, the CycleGAN works to output an image that resembles images from another (target) data domain. This model is trained using two loss functions: the standard adversarial objective of the original GAN (in order to generate meaningful images) and a ``cycle consistency'' loss that maximizes the domain similarity of the model's generated image space. The cycle consistency loss helps to guide the CycleGAN's generator to map the source image to the desired general features in the target domain. Later efforts related to CycleGAN introduced the identity loss \cite{cgan:id-loss-orig, cgan:id-loss-2nd}, which was further shown to stabilize the image-to-image translation process.

\textbf{Domain Adversarial Neural Network (DANN).} Another form of adversarial-based domain adaptation is based on the DANN model. The DANN \cite{dann} is a classification model that integrates an additional domain classifier that is also trained under an adversarial process. In essence, training a DANN is similar to training a GAN given that it has a class predictor which tries to maximize the accuracy of the prediction of two domains while another domain classifier tries to distinguish between the two domains based on the bottleneck latent representation produced by the system's encoder. However, in the DANN, we do not have access to the gradient for training the encoder in order to maximize the domain classifier loss. Therefore, a ``reverse gradient'' layer is used for the latent embedding output before going into the domain classifier layers. Mechanically, a reverse gradient layer is simply an identity function that operates within the backward propagation process (for computing parameter gradients) that further negates the resulting estimated gradient values.

\subsection{Metric Learning}
\label{sec:metric_learning}

The measurement of the distribution shift among datasets can provide useful information for closing the aforementioned reality gap, e.g., one can measure the domain shift by computing the maximum mean discrepancy (MMD) integral probability metric between two datasets~\cite{Gretton2012MMD}, minimizing this metric might potentially reduce the gap between datasets. To achieve this, normally, a function parameterized by a deep neural network is learned in a process known as metric learning. In the context of this work, we will utilize metric learning in our refinement process, e.g., such as with a Siamese Network \cite{siamese}, in order to quantitatively measure the degree of distribution shift between the synthetic samples we produce and the samples within real eye datasets.

\textbf{Siamese Network.} The Siamese Network \cite{siamese} consists of two identical deep neural networks that specifically share the same set of weights and structure (the structure and parameters are ``tied''). The purpose of the overall model is to output the representation of two different objects such that we may compute the distance between these objects in terms of their corresponding projections in a latent space, using a distance function such as the L2 (Euclidean) distance. Overall, learning in a Siamese Network is similar to learning an albeit complex distance function. Generally, a Siamese Network is trained to minimize a contrastive loss \cite{contrastive-loss-orig, contrastive-loss-2nd} which penalizes the system for any deviation in its predicted distance values from a chosen distance measurement. Other loss functions used for training include the triplet loss and the binary cross entropy loss \cite{siamese, siamese-loss, triplet-loss}.

\section{Methods}
\label{sec:methods}

\begin{figure}
    \centering
    \includegraphics[width=0.9\textwidth]{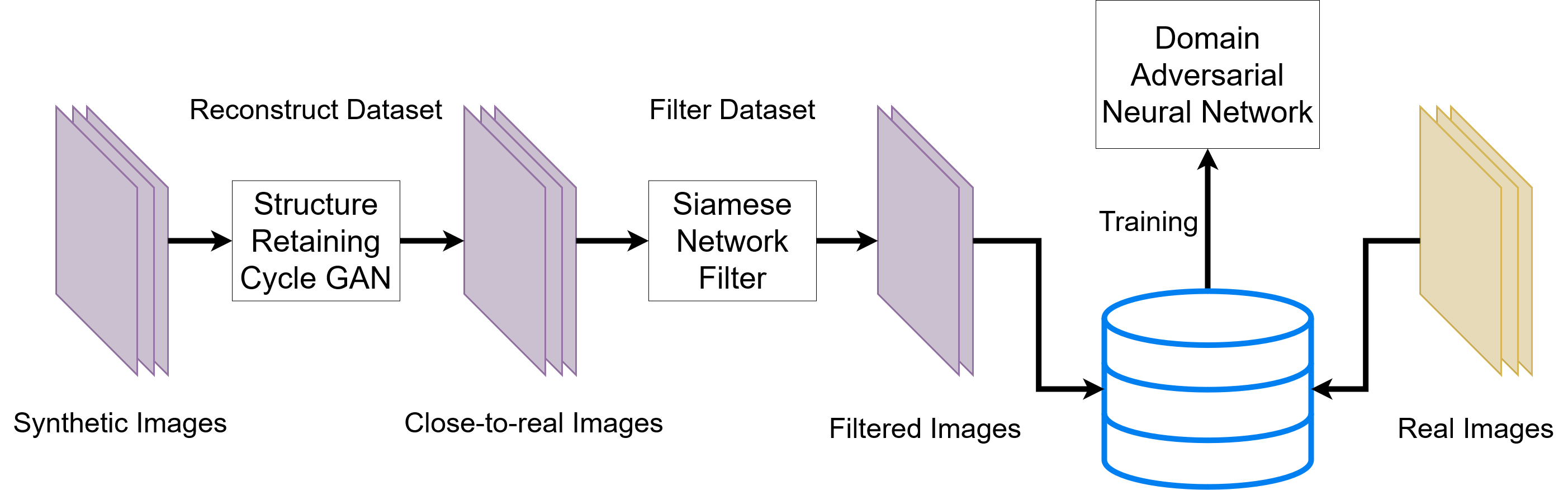}
    \caption{Overall process diagram of our proposed computational system for image segmentation. The synthetic images are first refined/processed using our novel Structure Retaining CycleGAN, then filtered by our Siamese Network that considers the distance between the latent representations of real and synthetic images, and finally placed into a training set that is used for training our adapted domain adversarial neural network.}
    \label{fig:overall_architecture}
\end{figure}

To address the \textit{sim2real} problem through domain adaptation, we develop and study the following: 
\textbf{1)} modifying the synthetic dataset such that it is closer to the real one so that the neural system benefits from the familiarity of the data at test time, and \textbf{2)} modifying the neural system to be capable of generalizing among different domains. Based on these notions, we propose a multi-step approach to the problem by utilizing image-to-image transfer (Section~\ref{subsec:cycle-gan-eye}), dataset filtering (Section~\ref{subsec:siamese-filtering}), and a domain generalization feature-based network (Section~\ref{subsec:dann-eye}). The overall architecture, as shown in Fig.~\ref{fig:overall_architecture}, first involves the implementation of our proposed  \textit{Structure Retaining CycleGAN}, which is a generalization of the CycleGAN~\cite{cycle-gan} that focuses on reconstructing the synthetic eye images under the constraint of matching the distribution of the real eye images. Next, we design a Siamese Network~\cite{siamese} for filtering out poorly-reconstructed images (i.e., a learned form of dataset pruning). Finally, we employ a model adapted from the domain-adversarial neural network structure~\cite{dann} which we will demonstrate has the ability to perform well across multiple domains.

\subsection{Structure Retaining CycleGAN} \label{subsec:cycle-gan-eye}

In the context of image generation/creation, a CycleGAN can be used to perform domain transfer such that images in the transferred domain are as diverse and as close to the target domain as possible~\cite{cycle-gan, cgan1, cgan2, cgan3}). In the context of domain adaptation, we also need to perform transferring within the label domain so as to ensure that the transferred labels match, i.e., the transferred pupil segmentation matches the exact pupil segmentation in the transferred eye image. The GeomaskGAN~\cite{geomask-gan} model uses a double input architecture which takes both the eye image and the segmentation label as input while performing the transfer. As an alternative, we propose another model that tries to preserve the structure of the eyes that avoids the need to transfer the segmentation label. We can achieve this by reconstructing images that have the same segmentation map structure as the original eye image. This inspiration comes from the intuition that there exists perceptually indistinguishable ``noise'', e.g., noise used to perturb an image in adversarial attack~\cite{adversarial-attack}, in the real data distribution, such that the model, when trained on the synthetic distribution, has reduced performance when inferred on this real distribution. This problem serves as the basis for a divergence in the pixel distribution between a synthetic image and a real one. 

 
\textbf{Problem Formulation.} Formally, we want to learn a function that maps an image from the synthetic eye domain $\mathcal{S} = \{\mathbf{s}^i\}^M_{i=1}$ to the real eye domain $\mathcal{R} = \{\mathbf{r}^i\}^N_{i=1}$ given the segmentation mask sets $\mathcal{M}_{\mathcal{S}} = \{\mathbf{m}_s^i\}^M_{i=1}$ and $\mathcal{M}_{\mathcal{R}} = \{\mathbf{m}_r^i\}^N_{i=1}$. We want to build the mapping function from domain $\mathcal{S}$ to domain $\mathcal{R}$ such that the transferred image $\mathbf{t}_{sr}^i \in \mathcal{T}_\mathcal{R}$ has the same original eye segmentation map $\mathbf{m}_s^i$ as the original image $\mathbf{s}^i$ and, furthermore, similar color features for each segmentation class $\mathbf{m}_r^i$ within the target domain image $\mathbf{r}^i$.




Similar to CycleGAN-like architectures~\cite{cycle-gan, cgan1, cgan2, cgan3}, our model architecture contains two separate image generators (under an encode-decoder setup) and two discriminators. Each generator maps an image from one domain to the other while each discriminator distinguishes the domain for each generated image. Particularly, we define two generators as $G_{\mathcal{S}\mathcal{R}}: \mathcal{S} \rightarrow \mathcal{T}_{\mathcal{R}} \approx \mathcal{R} $ and $G_{\mathcal{R}\mathcal{S}}: \mathcal{R} \rightarrow \mathcal{T}_{\mathcal{S}} \approx \mathcal{S}$, and we define two discriminators as $G_{\mathcal{S}}: \mathcal{T}_{\mathcal{S}} \cup \mathcal{S} \rightarrow \mathbb{R} = \{0,1\}$ and $G_{\mathcal{R}}: \mathcal{T}_{\mathcal{R}} \cup \mathcal{R} \rightarrow \mathbb{R} = \{0,1\}$ which output $0$ if the image sample comes from the generator and $1$ if the image sample comes from $\mathcal{S}$ or $\mathcal{R}$.

\textbf{Adversarial Loss.} In order to train the above model, we first employ the adversarial loss~\cite{gan}, similar to what has been done in the CycleGAN literature~\cite{cycle-gan}. Particularly, this objective function encourages the generator to produce images that are closer to the target domain while the discriminator must distinguish between images that actually come from the source distribution or those produced by the generator, i.e, $\text{min}_G\ \text{max}_D\ \mathcal{L}_{adv}$. The objective function for $G_{\mathcal{S}\mathcal{R}}$ and $D_{\mathcal{R}}$ is defined as:
\begin{equation}
\begin{aligned}
    \mathcal{L}_{adv} (G_{\mathcal{S}\mathcal{R}}, D_{\mathcal{R}}) =\ &\mathbb{E}_{\mathbf{r}\sim\mathcal{R}}[\log D_{\mathcal{R}}(\mathbf{r})] + \mathbb{E}_{\mathbf{s}\sim\mathcal{S}}[ 1 - \log{D_{\mathcal{R}}(G_{\mathcal{S}\mathcal{R}}(\mathbf{s}))} ] .
\end{aligned} 
\end{equation}

\textbf{Cycle Consistency Loss.} As mentioned before, this loss guides a network to learn a mapping function where recovered images are likely to closely match the original images~\cite{cycle-gan}. Particularly, we encourage the recovery of an image after translating it to another domain and back to the original domain. Similar to the original CycleGAN, we use the mean absolute error in order to compute the loss between the image before and after the domain transfer as follows:
\begin{equation}
\begin{aligned}
    \mathcal{L}_{cyc} (G_{\mathcal{S}\mathcal{R}}, G_{\mathcal{R}\mathcal{S}}) =\ &\mathbb{E}_{\mathbf{s}\sim\mathcal{S}}[\parallel G_{\mathcal{R}\mathcal{S}}(G_{\mathcal{S}\mathcal{R}}(\mathbf{s})) - \mathbf{s} \parallel_{1}] + \mathbb{E}_{\mathbf{r}\sim\mathcal{R}}[\parallel G_{\mathcal{S}\mathcal{R}}(G_{\mathcal{R}\mathcal{S}}(\mathbf{r})) - \mathbf{r} \parallel_{1}] .
\end{aligned}
\end{equation}


\textbf{Identity Loss.} This loss is often used in image-to-image translation problems in order to ensure that the color and tint of the translated image are as close to the original image as possible~\cite{cgan:id-loss-orig, cgan:id-loss-2nd}. Furthermore, the identity loss states that a generator for the target domain, when receiving an image from the same domain, must produce the image in the same domain, i.e, $G_{\mathcal{S}\mathcal{R}}(\mathbf{r}) \approx \mathbf{r}$. The identity loss, in our context, is formulated below as follows:
\begin{equation}
    \begin{aligned}
        \mathcal{L}_{id} (G_{\mathcal{S}\mathcal{R}}, G_{\mathcal{R}\mathcal{S}}) =\ &\mathbb{E}_{\mathbf{s}\sim\mathcal{S}}[\parallel G_{\mathcal{R}\mathcal{S}}(\mathbf{s}) - \mathbf{s} \parallel_{1}] + \mathbb{E}_{\mathbf{r}\sim\mathcal{R}}[\parallel G_{\mathcal{S}\mathcal{R}}(\mathbf{r}) - \mathbf{r} \parallel_{1}] .
    \end{aligned}
\end{equation}


\textbf{Edge Retaining Loss.} Note that the cycle consistency objective function does not guarantee that the transferred image in another (target) domain has the same segmentation structure (one can see an incorrect mapping in the center image of Fig.~\ref{fig:sample-dataset}). To overcome this problem, we propose that the structure of the image may be retained by keeping the edges of the image fixed throughout the translation process. In order to achieve this, we propose that the edge features of the original image should be as close as possible to the edge features of the translated image as well as the recovered image. For example, in the translation from domain $\mathcal{S}$ to domain $\mathcal{R}$, the edge features among the original image $\mathbf{s}\sim\mathcal{S}$, the translated image $G_{\mathcal{S}\mathcal{R}}(\mathbf{s})$, and the recovered image $G_{\mathcal{R}\mathcal{S}}(G_{\mathcal{S}\mathcal{R}}(\mathbf{s}))$ should be approximately equal. In support of this, we implemented the Sobel filter~\cite{sobel} (denoted as $\mathcal{F}$) in order to compute the edges of the image by performing a convolution over the designated image (the convolution operator is denoted as $*$). The objective function is then formulated as follows:
\begin{equation}\label{eqn:sr-cgan-loss-edge}
\begin{aligned}
    \mathcal{L}_{edge}(G_{\mathcal{S}\mathcal{R}}, G_{\mathcal{R}\mathcal{S}}) =\ &\mathbb{E}_{\mathbf{s}\sim\mathcal{S}} [\parallel\mathcal{F} * G_{\mathcal{S}\mathcal{R}}(\mathbf{s}) - \mathcal{F} * \mathbf{s} \parallel_{1} + \parallel\mathcal{F} * G_{\mathcal{S}\mathcal{R}}(\mathbf{s}) - \mathcal{F} * G_{\mathcal{R}\mathcal{S}}(G_{\mathcal{S}\mathcal{R}}(\mathbf{s})) \parallel_{1}]\\
    +\ &\mathbb{E}_{\mathbf{r}\sim\mathcal{R}} [\parallel\mathcal{F} * G_{\mathcal{R}\mathcal{S}}(\mathbf{r}) - \mathcal{F} * \mathbf{r} \parallel_{1} + \parallel\mathcal{F} * G_{\mathcal{R}\mathcal{S}}(\mathbf{r}) - \mathcal{F} * G_{\mathcal{S}\mathcal{R}}(G_{\mathcal{R}\mathcal{S}}(\mathbf{r})) \parallel_{1}] .
\end{aligned}
\end{equation}

\textbf{Color Mean and Variance Retaining Loss.} A generator that outputs the correct edge structure of the eye may not necessarily output the correct segmentation feature corresponding to its edges, e.g., the translated pupil is half-dark. Therefore, we propose a loss function that encourages the minimization of the statistical (mean and variance) color difference between the translated image and the target domain image, i.e., $\mathbf{t}_{sr}^i$ and $\mathbf{r}^i$, respectively. This loss works to increase the unity with respect to the color estimation within each eye part (e.g., pupil) when performing image translation. Particularly, we compute the difference in mean and variance for each corresponding segmentation class $k \in K$ number of classes, i.e., pupil, iris, sclera, and background, between image pairs, i.e., translated image $\mathbf{t}_{sr}^i$ and target domain image $\mathbf{r}^i$. Let the mean of the pixels for class $k$ of image $x$ be $\mu_k(x)$, where each class $k$ has $P_k$ number of pixels. As a result, we obtain the following:
\begin{equation}
\mu_k(x) = \frac{1}{P_k}\sum_{p=1}^{P_k}{x_p} .
\end{equation}
Given the above, the color mean retaining loss function is then represented as follows:
\begin{equation}
\begin{aligned}
    \mathcal{L}_{mean}(G_{\mathcal{S}\mathcal{R}}, G_{\mathcal{R}\mathcal{S}}) =\ \mathbb{E}_{\mathbf{s}\sim\mathcal{S},\mathbf{r}\sim\mathcal{R}} \sum_k{|\mu_k(G_{\mathcal{S}\mathcal{R}}(\mathbf{s})) - \mu_k(\mathbf{r}) |} + \mathbb{E}_{\mathbf{r}\sim\mathcal{R},\mathbf{s}\sim\mathcal{S}} \sum_k{ |
\mu_k(G_{\mathcal{R}\mathcal{S}}(\mathbf{r})) - \mu_k(\mathbf{s}) |} .
\end{aligned}
\end{equation}
Similarly, let the variance of the pixels for class $k$ of image $x$ be $\sigma_k(x)$, where each class $k$ has $P_k$ number of pixels. We then compute the following:
\begin{equation}
\sigma_k(x) = \frac{1}{P_k}\sum_{p=1}^{P_k}{(x_p - \mu_k(x))^2} .
\end{equation}
The color variance retaining loss function is then represented finally in the following manner:
\begin{equation}
\begin{aligned}
   \mathcal{L}_{var}(G_{\mathcal{S}\mathcal{R}}, G_{\mathcal{R}\mathcal{S}}) = \mathbb{E}_{\mathbf{s}\sim\mathcal{S},\mathbf{r}\sim\mathcal{R}} \sum_k{ | \sigma_k(G_{\mathcal{S}\mathcal{R}}(\mathbf{s})) - \sigma_k(\mathbf{r}) | } + \mathbb{E}_{\mathbf{r}\sim\mathcal{R},\mathbf{s}\sim\mathcal{S}} \sum_k{|
\sigma_k(G_{\mathcal{R}\mathcal{S}}(\mathbf{r})) - \sigma_k(\mathbf{s}) |} .
\end{aligned}
\end{equation}

\textbf{Final Structure Retaining CycleGAN Objective Function.} Given the above designed set of loss functions, the full objective function used to train our neural system is the following:
\begin{equation}
\begin{aligned}
    \mathcal{L}(G_{\mathcal{S}\mathcal{R}}, G_{\mathcal{R}\mathcal{S}}, D_{\mathcal{R}}, D_{\mathcal{S}}) =\  &\mathcal{L}_{adv}(G_{\mathcal{S}\mathcal{R}}, D_{\mathcal{R}}) + \mathcal{L}_{adv}(G_{\mathcal{R}\mathcal{S}}, D_{\mathcal{S}}) 
    + \gamma_{cyc} \mathcal{L}_{cyc}(G_{\mathcal{S}\mathcal{R}}, G_{\mathcal{R}\mathcal{S}}) 
    + \gamma_{id} \mathcal{L}_{id}(G_{\mathcal{S}\mathcal{R}}, G_{\mathcal{R}\mathcal{S}}) \\
    +\ &\gamma_{edge} \mathcal{L}_{edge}(G_{\mathcal{S}\mathcal{R}}, G_{\mathcal{R}\mathcal{S}}) 
    + \gamma_{mean} \mathcal{L}_{mean}(G_{\mathcal{S}\mathcal{R}}, G_{\mathcal{R}\mathcal{S}}) 
    + \gamma_{var} \mathcal{L}_{var}(G_{\mathcal{S}\mathcal{R}}, G_{\mathcal{R}\mathcal{S}}) \\
\end{aligned}
\end{equation}
where $\gamma_{cyc}$, $\gamma_{id}$, $\gamma_{edge}$, $\gamma_{mean}$, $\gamma_{var}$ are the coefficients that control the effect that each corresponding loss term has on the full system optimization process.


\subsection{Siamese Network Filtering}\label{subsec:siamese-filtering}
After a synthetic image has been reconstructed to be closer to the real eye image distribution, there will still exist parts of the images that are not very close to the real distribution. As can be seen from the PCA plot of intermediate latent vectors (see Fig.~\ref{fig:pca-dann-srcgan-real}), the real image distribution does not fully cover the reconstructed image distribution. In order to overcome this, we remove images that are far away from the real image distribution by thresholding their distance to the real image dataset's centroid. In order to measure the distance of one image from the other, we craft a Siamese Network~\cite{siamese, sia-domain-adapt, sia-motion-dark} that infers the latent representation of each image such that the distance, i.e, the L2 distance, between images from two different domains should be far from each other. As a result, we employ a contrastive loss~\cite{contrastive-loss-orig, contrastive-loss-2nd} to achieve this goal.


\textbf{Problem Formulation.} Concretely, we construct a Siamese Network that maps from image/pixel space to a latent vector of size $n$ ($n=2$ in our case): $f: \mathbb{I} \rightarrow \mathbb{R}^2$. We then filter the reconstructed dataset $\mathcal{T}_\mathcal{R}$ by thresholding each synthetic image's distance-to-centroid on the real dataset. In particular, we first compute the centroid vector representation $c_{\mathcal{R}}$ of the real dataset in the following way:
\begin{equation} \label{eqn:sia-centroid-repr}
\begin{aligned}
    c_{\mathcal{R}} = \mathbb{E}_{\mathbf{r}\sim\mathcal{R}} \left[ f(\mathbf{r}) \right] .
\end{aligned}
\end{equation}
We next compute the distance $d^i$ of each reconstructed image $\mathbf{t}_{sr}^i \in \mathcal{T}_\mathcal{R}$ to the real domain's centroid:
\begin{equation}
\begin{aligned}
    d^i = \parallel f(\mathbf{t}_{sr}^i) - c_{\mathcal{R}} \parallel^2_2.
\end{aligned}
\end{equation}
Finally, we may then choose only images in which the distance $d^i$ is smaller than a certain threshold ($0.005$) to ultimately synthesize a filtered dataset.


\subsection{A Domain-Adversarial Neural Network (DANN) for Segmentation}\label{subsec:dann-eye}


Current segmentation methodologies have excelled in working with domain-specific datasets. However, when it comes to carrying out inference over different domains, performance degradation is often observed, i.e., training a segmentation model on a synthetic dataset yields low mean intersection over union on the test dataset of real images. This problem stems from the fact that the images from both datasets essentially come from two different distributions (as mentioned before, this violates the typical assumption that the data constituting the training and test sets are identically distributed).

One way to close the above domain gap is to learn a feature extractor that can generalize across different domains in its latent space. To achieve this, we took inspiration from the training of domain-adversarial neural network (DANN)~\cite{dann}. In our problem context, we propose constructing a decoder head (for segmentation) instead of a class predictor as in the original DANN. Our goal is to make the encoder's output the feature/component that generalizes across two domains. As a result, we need to reverse the gradients \cite{dann} that come from the domain classifier. In particular, given a set of $n$-dimensional latent vectors $\mathbb{R}^n$, the set of $2$D images (height $h$, width $w$, $c$ channels) $\mathbb{I}^{h\times w\times c}$, $K$ number of segmentation classes, the encoder (of image $x$) $e_{\theta}(x): \mathbb{I}^{h\times w\times c} \rightarrow \mathbb{R}^n$ (which has $5$ down blocks \cite{ritnet}), the decoder $d_{\theta}(e(x)): \mathbb{R}^n \rightarrow \mathbb{I}^{h\times w\times K}$ (which has $4$ up blocks \cite{ritnet}), and the domain classifier $f_{\lambda}(e(x)): \mathbb{R}^n \rightarrow \mathbb{R}^1$, the optimization objective of our network is formally the following:
\begin{equation}
\begin{aligned}
    \underset{e_{\theta}, d_{\theta}, f_{\lambda}}{\min}\ \underset{e_{\theta}}{\max}\ \mathcal{L} (e_{\theta}, d_{\theta}, f_{\lambda})
    &\Leftrightarrow \underset{e_{\theta}, d_{\theta}}{\min}\ \mathcal{L}_{\text{ritnet}}(e_{\theta}, d_{\theta}) \quad \text{and} \quad \underset{e_{\theta}}{\max}\ \underset{f_{\lambda}}{\min}\ \mathcal{L}_{\text{domain}} (e_{\theta}, f_{\lambda})\\
    \text{with } \mathcal{L}_{\text{domain}} (e_{\theta}, f_{\lambda}) &= - \mathbb{E}_{x \sim \mathbb{I}^{h\times w\times c}} \left[ l \log{(f_{\lambda}(e_{\theta}(x)))} + (1-l) \log{(1 - f_{\lambda}(e_{\theta}(x)))} \right] .
\end{aligned}
\end{equation}
Note that the above objective function can be deconstructed into two key components. First, it involves optimizing the segmentation model (RITnet) loss function, which further decomposes into the minimizing of a generalized dice loss \cite{ritnet, gdl1, gdl-2}, a boundary-aware loss \cite{ritnet}, and a surface loss \cite{surface-loss} explained in Section~\ref{sec:lit_review}. Second, our domain classifier loss $\mathcal{L}_{\text{domain}} (e_{\theta}, f_{\lambda})$ can be expressed as the binary cross-entropy loss between the predicted dataset classification of image $x$ and its label $l$ -- where $l$ denotes which domain that the image $x$ comes from. Our objective is to minimize this domain classifier loss with respect to the domain classifier $f_{\lambda}$ such that it maximizes the same loss for the encoder $e_{\theta}$. This also means that the gradient signal that optimizes the domain classifier $f_\lambda$ should be negated when it flows through the encoder $e_\theta$. To achieve this, we utilize the reverse gradient layer~\cite{dann} at the end of the encoder so as to make sure that the gradients minimize the domain classifier loss for the domain classifier $f_{\lambda}$ while still ensuring that the encoder $e_{\theta}$ is maximizing that same loss value.

\section{Results and Discussion}
\label{sec:results}


\begin{figure}[h]
    \centering
    \begin{subfigure}[b]{0.17\textwidth}
        \includegraphics[width=1\textwidth]{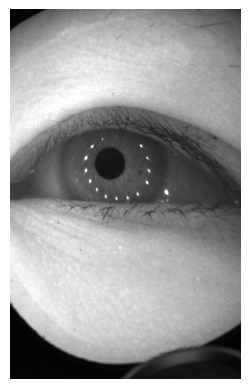}  
    \end{subfigure}
    \begin{subfigure}[b]{0.17\textwidth}
        \includegraphics[width=1\textwidth]{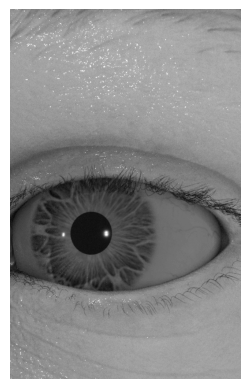}  
    \end{subfigure}
    \begin{subfigure}[b]{0.17\textwidth}
        \includegraphics[width=1\textwidth]{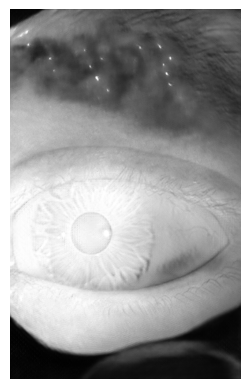}  
    \end{subfigure}
    \begin{subfigure}[b]{0.17\textwidth}
        \includegraphics[width=1\textwidth]{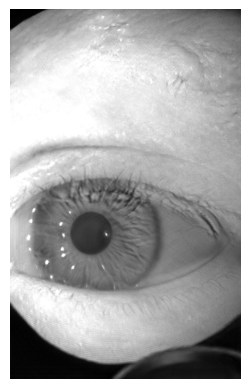}  
    \end{subfigure}
    \begin{subfigure}[b]{0.17\textwidth}
        \includegraphics[width=1\textwidth]{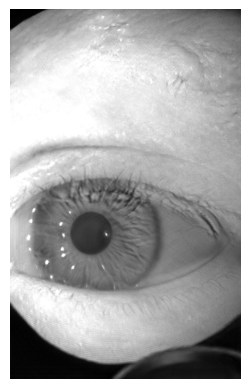}  
    \end{subfigure}
    \hfill
    \caption{Sample images from datasets used in our experiments. From left to right: OpenEDS (target domain), and four synthetic/constructed source domains - RITEyes, CGAN, SRCGAN, and SRCGAN-S.} \label{fig:sample-dataset}
\end{figure}

\begin{table}[ht]
    \parbox{.53\linewidth}{
        \centering
        \begin{tabular}{l||lll}
        \diagbox{Set}{Domain} & \textbf{Source} & \textbf{Filtered Source} & \textbf{Target} \\ \hline \hline
        \textbf{Train} & 9,216 & 2,915 & 8,916 \\ 
        \textbf{Validation} & 1,024 & 323 & 2,403  \\ \hline \hline
        \textbf{Total} & 10,240 & 3,238 & 11,319 \\ 
    \end{tabular}
    \caption{Number of images used for each dataset. The ``Source'' domains include RITEyes, CGAN, and SRCGAN, the ``Filtered Source'' domain includes SRCGAN-S, and the ``Target'' domain includes OpenEDS.} \label{tab:num-images} 
    }
    \hfill
    \parbox{.44\linewidth}{
        \centering
        \begin{tabular}{c||ccccc}
        \diagbox{N}{M} & \textbf{64} & \textbf{256} & \textbf{1,024} & \textbf{2,048} & \textbf{4,096} \\ \hline \hline
        \textbf{0} & 1,600 & 800 & 200 & 100 & 70 \\ 
        \textbf{64} & 400 & 150 & 120 & 100 & 70 \\ 
        \textbf{8,192} & 120 & 100 & 80 & 70 & 60 \\ 
    \end{tabular}
    \caption{Number of training epochs for each combination of $M$ images in the source domain and $N$ images in target domain.}
    \label{tab:training-epochs}
    }
\end{table}


\textbf{Dataset Details.} Fig.~\ref{fig:sample-dataset} shows sample images from five datasets. The first image is from the real OpenEDS dataset ~\cite{openeds}, the second image is synthetic and was generated using the RITEyes rendering pipeline~\cite{riteyes}, and we have constructed the remaining three datasets - CGAN (created using the CycleGAN method \cite{cycle-gan} described in related work, SRCGAN (created using our Structure Retaining CycleGAN method described in Section \ref{subsec:cycle-gan-eye}, and SRCGAN-S (created by filtering the SRCGAN dataset through our Siamese Network described in Section \ref{subsec:siamese-filtering}.
We denote the OpenEDS dataset \cite{openeds} as the target dataset/domain that the other datasets must be adapted to (the other four are labeled as the source datasets). Both the source (synthetic) and target (real) dataset have four types of label - pupil, sclera, iris, and background - although they may have different locations within the images. The image resolution of the Open EDS dataset is $400 \times 640$ pixels~\cite{openeds}, so we used the same resolution  for the synthetic dataset generated by the RITEyes pipeline \cite{riteyes} as well as for the three constructed datasets (CGAN, SRCGAN, and SRCGAN-S) in order to reduce variance in our (deep) neural networks' input space. We also generated synthetic images with the same number of channels (grayscale images) and the same number of segmentation classes, i.e., pupil, iris, sclera, background. Visual inspection suggests that even the more convincing of artificial images differ from the real images along several dimensions, including the realism in eye lashes, skin texture, and iris texture. The training procedure meant to reduce these differences proceeded under a 3-fold cross-validation scheme. The number of images used in each dataset is shown in Table \ref{tab:num-images}. While generating data for training, the data augmentation methods described in RITnet \cite{ritnet} were used. These include vertical axis reflection, Gaussian blur (with a kernel of size $7 \times 7$ with standard deviation $2 \le \sigma \le 7$), image translation of $0$-$20$ pixels along both axes, image corruption by drawing $2$-$9$ random vertical and horizontal thin lines, and image corruption using a starburst pattern. Each of the augmentation methods has a probability of $0.2$ of being selected when generating training images. The number of training epochs/iterations is shown in Table~\ref{tab:training-epochs}. Note that the number of epochs is manually adjusted to be higher while training datasets with lower number of images. This is done to relatively balance the total training steps across training instances.


\textbf{Architecture of CycleGAN-based Models.} We utilize elements of the ResNet architecture \cite{resnet, cycle-gan} within our CycleGAN-based models (i.e. CycleGAN and Structure Retaining CycleGAN). The generator consists of a convolution neural network (CNN) block, followed by downsampling by a factor of $4$, which is then followed by $8$ residual blocks \cite{resnet}, and finally, upsampling is applied with a factor of $4$ to obtain the generated image. For the discriminator, we use $4$ CNN blocks with a stride of $4$ and a leaky ReLU activation function ($\alpha = 0.2$), followed by a linear transformation layer that outputs a single neuron representing the prediction of whether an input is real or fake.

In Section~\ref{subsec:cycle-gan-eye}, a number of hyperparameters that control the Structure Retaining CycleGAN objective function are mentioned. We choose to keep the cycle loss $\gamma_{\text{cyc}}$ and identity loss $\gamma_{\text{id}}$ coefficient the same as in prior work \cite{cycle-gan, cgan:id-loss-orig, cgan:id-loss-2nd} (i.e, $10$). For the coefficients of the newly proposed objective functions ($\gamma_{\text{edge}}$, $\gamma_{\text{mean}}$, $\gamma_{\text{var}}$), we perform a test over multiple combinations of parameters and choose the best combination based on model performance, i.e., with respect to mean distance to the real distribution's centroid. The best combination of hyperparameters was $\gamma_{\text{edge}} = 0.1$, $\gamma_{\text{mean}} = 0$, $\gamma_{\text{var}} = 60$.

\begin{figure}[!ht]
    \centering
    \begin{subfigure}[b]{0.28\textwidth}
        \includegraphics[width=1\textwidth]{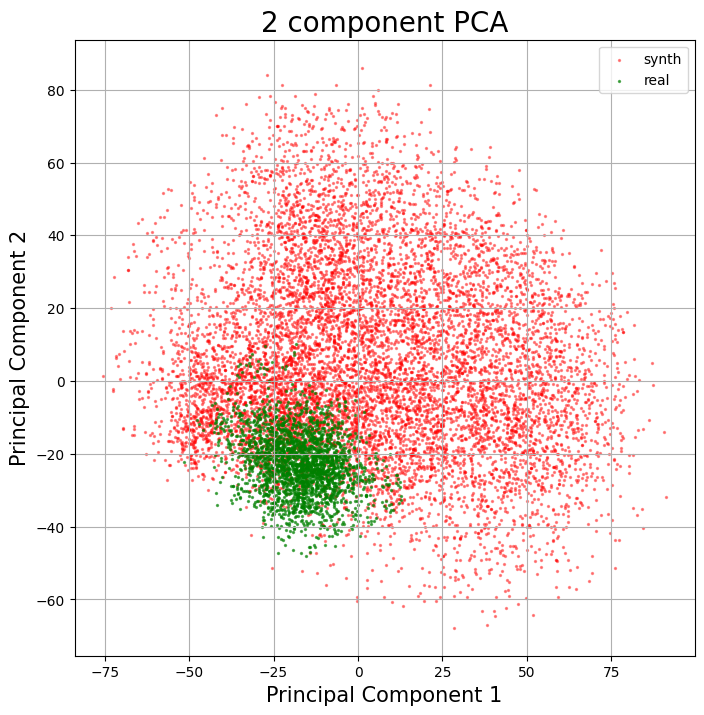}  
    \end{subfigure}
    \hfill
    \begin{subfigure}[b]{0.28\textwidth}
        \includegraphics[width=1\textwidth]{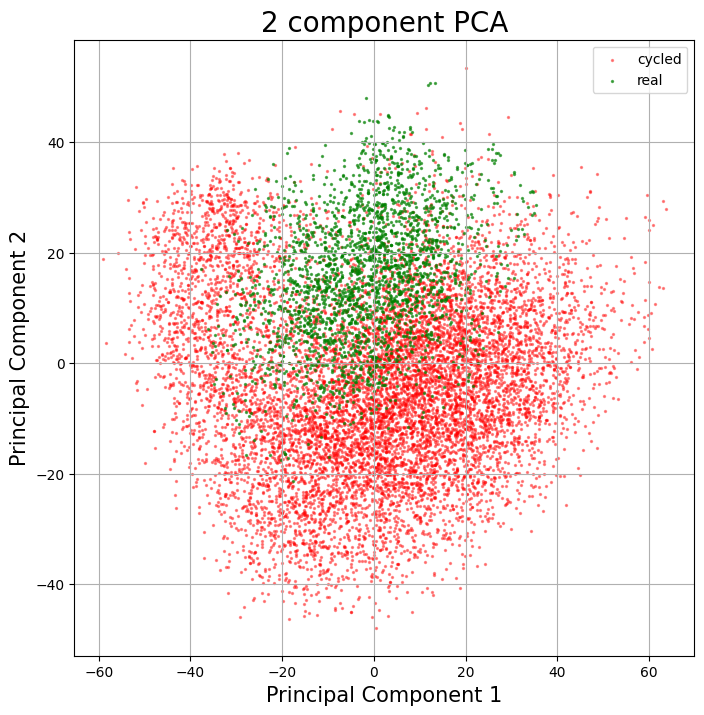}  
    \end{subfigure}
    \hfill
    \begin{subfigure}[b]{0.28\textwidth}
        \includegraphics[width=1\textwidth]{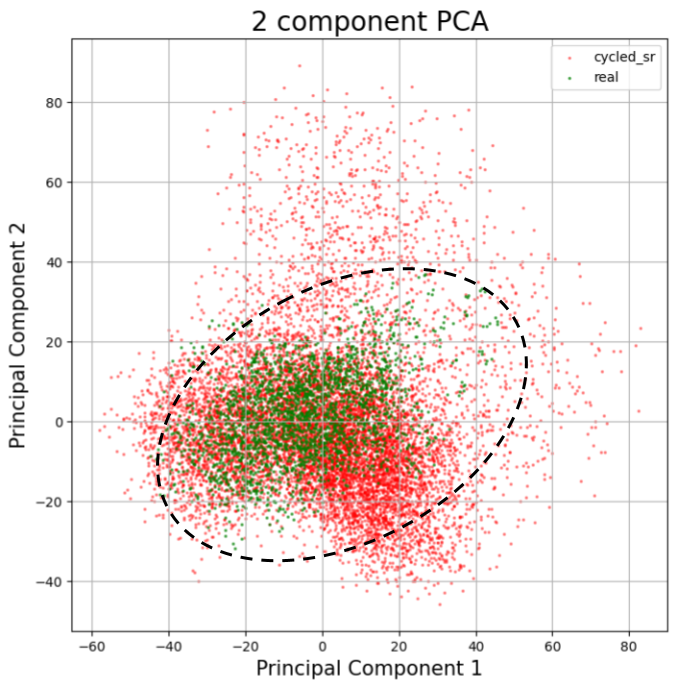}  
    \end{subfigure}
    \caption{Comparison of PCA plots of intermediate latent vectors for source and target domains produced by the DANN module (described in Section~\ref{subsec:dann-eye}). Left: RITEyes (red) vs. OpenEDS (green). Middle: CGAN (red) vs. OpenEDS (green). Right: SRCGAN (red) vs. OpenEDS (green). Note that red dots inside the ellipse make up the SRCGAN-S distribution which represents filtered images that are close to the real distribution.}
    \label{fig:pca-dann-srcgan-real}
\end{figure}

There are multiple ways to measure the performance of CycleGAN-based models such as Inception Score where generated images are evaluated based on predictability and diversity~\cite{cgan1, cgan2}. In our context, the CycleGAN-based model-generated images have to be both meaningful, i.e., predictable by the Inception Network classifier, as well as meet the goal of being close to the real domain. We choose to utilize mean intersection over union (mIoU) to compare classification performance across models in the context of segmentation prediction. In addition, we measure the closeness of generated datasets to the real dataset. In order to achieve this, we first employ the Inception Network~\cite{inceptionv4} used in the Siamese Network (Section~\ref{subsec:siamese-filtering}) to infer the latent representation vector corresponding to each image. Then, we compute the statistics for each vector in the source domain distribution compared to the centroid of the real distribution, i.e., using the L2 distance. We next compute the real distribution's centroid by averaging every real image vector. The mean distance-to-real-centroid formula can then be formally stated in the following manner:
\begin{equation}
\begin{aligned}
    \mu_d^{\mathcal{C}} = \mathbb{E}_{c\sim \mathcal{C}} \left[ \parallel f(c) - c_{\mathcal{R}} \parallel^2_2 \right] .
\end{aligned}
\end{equation}
where $\mathcal{C}$ represents the source dataset being evaluated and $c_{\mathcal{R}}$ is the vector representation of the real dataset's centroid as calculated in Equation~\ref{eqn:sia-centroid-repr}. 
We obtain $\mu_d^{\text{CGAN}} \approx 0.023$ and $\mu_d^{\text{SRCGAN}} \approx 0.005$ (the best given the hyperparameter combination). 
The results notably align with the PCA plots of the DANN module (see Section \ref{subsec:dann-eye}) across different datasets as shown in Fig.~\ref{fig:pca-dann-srcgan-real}.

\begin{figure}[h]
    \centering
    \begin{tikzpicture}
        \node[anchor=south west,inner sep=0] (image) at (0,0) {
            \begin{subfigure}[b]{0.45\textwidth}
                \includegraphics[width=1\textwidth]{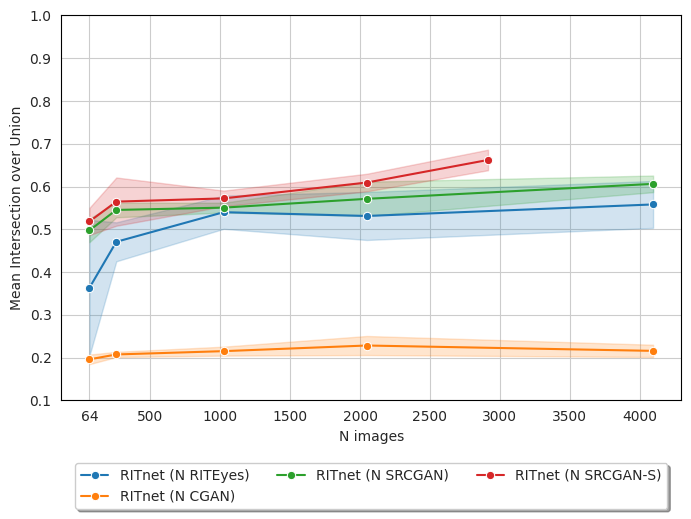}  
            \end{subfigure}
            \begin{subfigure}[b]{0.45\textwidth}
                \includegraphics[width=1\textwidth]{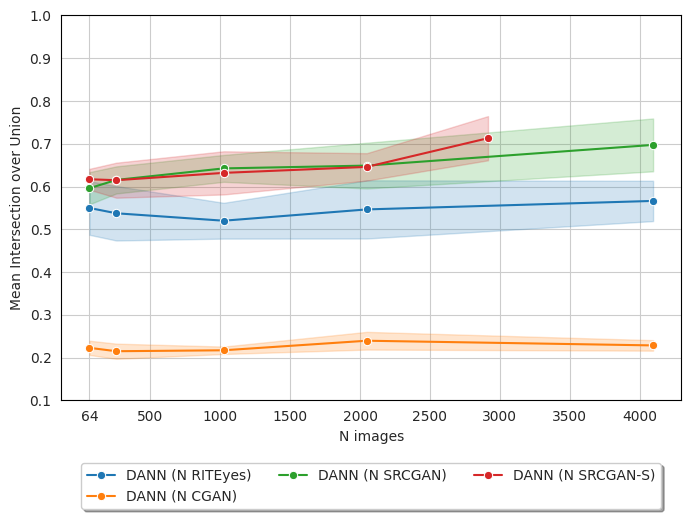}  
            \end{subfigure}
        };
    \end{tikzpicture}
    \caption{Model performance comparison on the real target dataset (OpenEDS) of the RITnet segmentation network (Left) and our DANN segmentation network (Right). Both models were trained on the $4$ source domains (see Figure \ref{fig:sample-dataset}). Shaded regions depict $\pm 1$ standard deviation in the $3$-fold cross validation scheme.} 
    \label{fig:srcgan-miou}
\end{figure}

\begin{table}
  \centering
  \subfloat[RITnet]{%
    \begin{tabular}{c||cccccc||c}
        \diagbox{Dataset}{N} & \textbf{64} & \textbf{256} & \textbf{1,024} & \textbf{2,048} & \textbf{2,915} & \textbf{4,096} & \textbf{$\text{mmIoU}^{\mathcal{C}}$} \\ \hline \hline
        RITEyes & 0.36$\pm$0.11 & 0.47$\pm$0.03 & 0.54$\pm$0.03 & 0.53$\pm$0.04 & - & 0.56$\pm$0.04 & 0.49$\pm$0.09 \\ 
        CGAN & 0.20$\pm$0.01 & 0.21$\pm$0.00 & 0.22$\pm$0.01 & 0.23$\pm$0.02 & - & 0.22$\pm$0.01 & 0.21$\pm$0.01 \\ 
        SRCGAN (ours) & 0.50$\pm$0.02 & 0.55$\pm$0.01 & 0.55$\pm$0.01 & 0.57$\pm$0.03 & - & \cellcolor{LightCyan}\textbf{0.61$\pm$0.01} & 0.55$\pm$0.04 \\ 
        SRCGAN-S (ours) & \cellcolor{LightCyan}\textbf{0.52$\pm$0.02} & \cellcolor{LightCyan}\textbf{0.56$\pm$0.04} & \cellcolor{LightCyan}\textbf{0.57$\pm$0.01} & \cellcolor{LightCyan}\textbf{0.61$\pm$0.01} & \cellcolor{LightCyan}\textbf{0.66$\pm$0.02} & - & \cellcolor{LightCyan}\textbf{0.59$\pm$0.05} \\ \hline
    \end{tabular}
  }
  \vspace{.3cm}
  \subfloat[DANN]{%
    \begin{tabular}{c||cccccc||c}
        \diagbox{Dataset}{N} & \textbf{64} & \textbf{256} & \textbf{1,024} & \textbf{2,048} & \textbf{2,915} & \textbf{4,096} & \textbf{$\text{mmIoU}^{\mathcal{C}}$} \\ \hline \hline
        RITEyes & 0.55$\pm$0.04 & 0.54$\pm$0.05 & 0.52$\pm$0.03 & 0.55$\pm$0.05 & - & 0.57$\pm$0.03 & 0.54$\pm$0.04 \\ 
        CGAN & 0.22$\pm$0.01 & 0.22$\pm$0.01 & 0.22$\pm$0.01 & 0.24$\pm$0.01 & - & 0.23$\pm$0.01 & 0.22$\pm$0.01 \\ 
        SRCGAN (ours) & 0.60$\pm$0.03 & \cellcolor{LightCyan}\textbf{0.62$\pm$0.02} & \cellcolor{LightCyan}\textbf{0.64$\pm$0.02} & \cellcolor{LightCyan}\textbf{0.65$\pm$0.04} & - & \cellcolor{LightCyan}\textbf{0.70$\pm$0.04} & \cellcolor{LightCyan}\textbf{0.64$\pm$0.04} \\ 
        SRCGAN-S (ours) & \cellcolor{LightCyan}\textbf{0.62$\pm$0.02} & 0.61$\pm$0.03 & 0.63$\pm$0.04 & \cellcolor{LightCyan}\textbf{0.65$\pm$0.02} & \cellcolor{LightCyan}\textbf{0.71$\pm$0.04} & - & \cellcolor{LightCyan}\textbf{0.64$\pm$0.04} \\
    \end{tabular}
  }
  \vspace{.2cm}
  \caption{Model performance (mIoU and $\text{mmIoU}^\mathcal{C}$) comparison on the real target dataset (OpenEDS) of the RITnet segmentation network (a) and our DANN segmentation network (b). Both models were trained on different number of images ($N$) from the $4$ source domains (see Fig.  \ref{fig:sample-dataset}). The final standard deviation of $\text{mmIoU}^\mathcal{C}$ is computed based on Bessel's correction formula. Best performance highlighted in bold text cyan cell.} \label{fig:srcgan-miou2}
\end{table}

The mIoU results of our system, trained on each source dataset, also demonstrates that using the Structure Retaining CycleGAN architecture improves the performance of the model when processing real datasets (Fig.~\ref{fig:srcgan-miou} and Table~\ref{fig:srcgan-miou2}). Specifically, 
let $\{\mathcal{C}^N\}$ be the set of all dataset $\mathcal{C}$ instances that have $N$ number of training images, e.g., for $\{\text{SRCGAN-S}^N\}$, $N \in \{64, 128, 1024, 2048, 2915\}$. We may then compute the overall representative mean of the individual mIoUs ($\text{mmIoU}^\mathcal{C}$) for each dataset $\mathcal{C}$ as: 
\begin{equation}
\begin{aligned}
    \text{mmIoU}^\mathcal{C} = \mathbb{E}_{\mathcal{C}_i \in \{\mathcal{C}^N\}}{\text{mIoU}^{\mathcal{C}_i}}
\end{aligned}
\label{eqn:mmiou}
\end{equation}
where $\text{mIoU}^{\mathcal{C}_i}$ is the mIoU of the segmentation model trained on the $i$-th instance in that collection of instances. In other words, we average all segmentation mIoU scores for a model over every dataset $\mathcal{C}$ instance for a single dataset $\mathcal{C}$ in order to retrieve the necessary statistics for each dataset $\text{mmIoU}^\mathcal{C}$. We observe that the RITnet segmentation network, when trained with ($\mathcal{C}=$) SRCGAN-S images, has $0.04, 0.38, \text{ and } 0.10$ higher $\text{mmIoU}^\mathcal{C}$ measurements than when trained with SRCGAN, CGAN, and RITEyes images, respectively. Similarly, the DANN model, when trained with ($\mathcal{C}=$) SRCGAN-S images, yields $0.00, 0.42, \text{ and } 0.10$ higher $\text{mmIoU}^\mathcal{C}$ scores as compared to training with SRCGAN, CGAN, RITEyes images, respectively (see Table~\ref{fig:srcgan-miou2}). Note that, in Fig.~\ref{fig:srcgan-miou} and Table~\ref{fig:srcgan-miou2}, there is no data for the  number of images greater than $2{\small,}915$ for SRCGAN-S dataset because there are only $2{\small,}915$ images in the training set (see Table~\ref{tab:num-images}). This result shows that we have developed a dataset production and refinement method that maps from the synthetic domain to the real domain, improving the plausibility of the synthetic images and crucially providing a means of closing the domain/\textit{sim2real} gap.



\begin{figure}[h]
    \centering
    \begin{subfigure}[b]{0.4\textwidth}
        \includegraphics[width=1\textwidth]{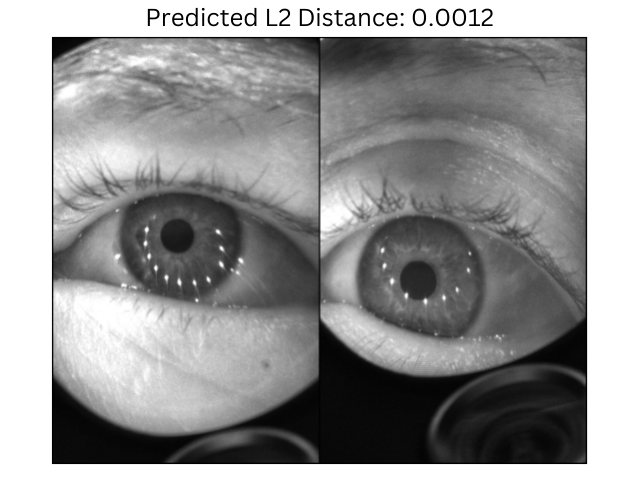}  
    \end{subfigure}
    \begin{subfigure}[b]{0.4\textwidth}
      \includegraphics[width=1\textwidth]{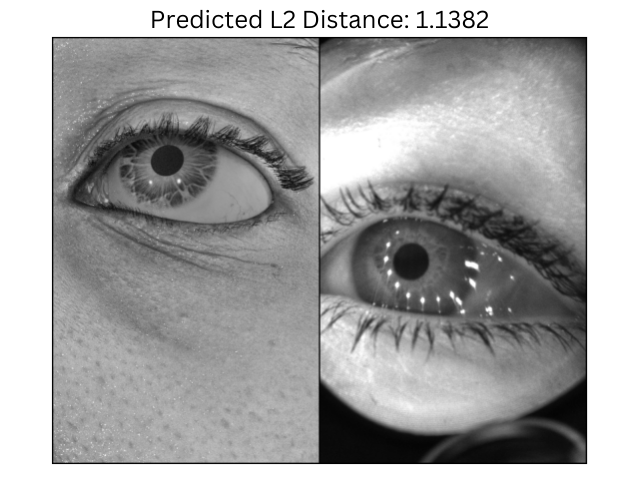}  
    \end{subfigure}
    \caption{Sample distance prediction measurements of the Siamese Network for  two images from OpenEDS dataset (left) and an image from RITEyes and one from OpenEDS respectively (right).} \label{fig:sia-sample-pred}
\end{figure}

\textbf{Siamese Network for Image Filtering.} We train a Siamese Network based on the Inceptionv4 architecture~\cite{inceptionv4} as the feature extractor, resulting in a total number of $27{\small,}465{\small,}826$ parameters. Concretely, we train the Siamese Network for $20$ epochs with $10{\small,}000$ pairs of the same source, same target, and different domain images for each epoch. We achieve a (final) $0.0001$ contrastive loss when estimating the L2/Euclidean distance between image pairs given that the labeled margin between two different-domain images is one. This means that the images that are close to each other (within the same domain) will have an estimated distance close to zero while images that are further away from one another (from different domains) will have a distance around one. Note that this can be observed from the sample estimation of L2 distance as shown in Fig.~\ref{fig:sia-sample-pred}.

After training the Siamese Network, the resulting model is used to output a latent representation of each image which is then used to filter the SRCGAN dataset. Through experimentation, we set the distance threshold used for filtering to be $0.005$ (the best mean distance-to-real-centroid $\mu_d^\text{SRCGAN}$ considering datasets generated from various Structure Retaining CycleGAN models). We then train all of the models on this filtered dataset and measure the mIoU. As seen in Fig.~\ref{fig:srcgan-miou} and Table~\ref{fig:srcgan-miou2}, the model trained on the filtered dataset desirably results in a higher mIoU compared to every other model. This score is specifically higher than that of the model trained on the non-filtered synthetic dataset (RITEyes) by about 10\%. This result shows that, after processing the synthetic dataset with our pipeline's other modules, we can further refine this dataset by filtering out images that are estimated not close to the real image distribution. This, in effect, further boosts the performance of the eye segmentation network.

\textbf{Domain Adversarial Neural Network.} Our DANN module consists of a segmentation network based on RITnet with the addition of a domain classifier. The domain classifier is a fully-connected neural network. The bottleneck output of the segmentation network, i.e., encoder output, is then fed into a reverse gradient layer. This reverse gradient layer acts as the identity function in the forward pass and negation function in the backward pass. The domain classifier then takes this in as input and outputs a single classification probability (score). The network classifier is made up of five dense layers that use ReLU activation functions except for the last layer, which makes use of the logistic sigmoid activation.

\begin{figure}[h]
    \begin{tikzpicture}
        \node[anchor=south west,inner sep=0] (image) at (0,0) {
            \begin{subfigure}[b]{0.33\textwidth}
                \includegraphics[width=1\textwidth]{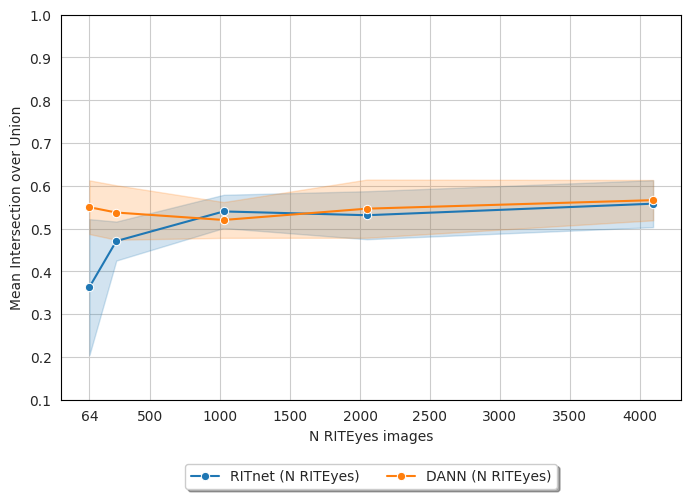}  
            \end{subfigure}
            \hfill
            \begin{subfigure}[b]{0.33\textwidth}
                \includegraphics[width=1\textwidth]{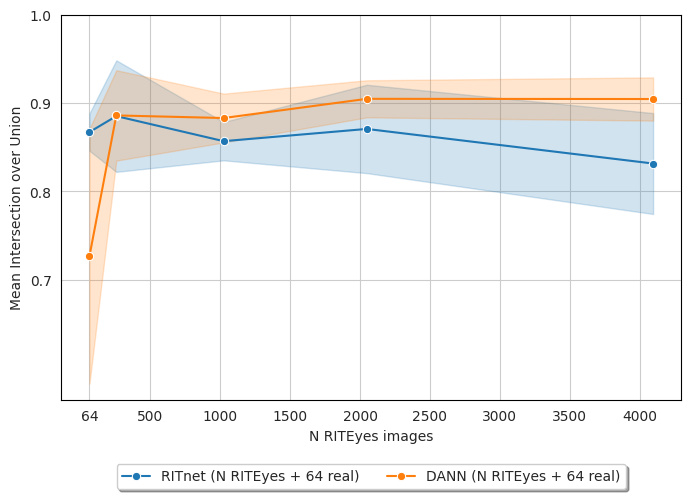}  
            \end{subfigure}
            \hfill
            \begin{subfigure}[b]{0.33\textwidth}
              \includegraphics[width=1\textwidth]{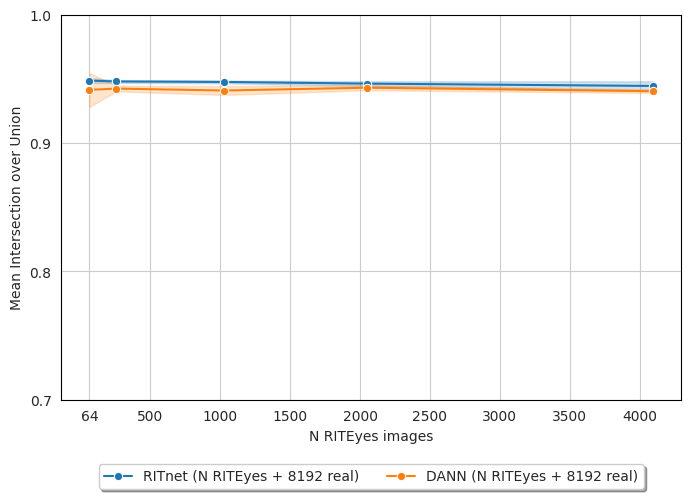}  
            \end{subfigure}
        };
    \end{tikzpicture}
    \caption{Performance comparison of DANN (orange) and RITnet (blue) when training on RITEyes dataset together with $0$ (left), $64$ (middle), or $8192$ (right) images from the OpenEDS dataset.} \label{fig:ritnet-dann-compare}
\end{figure}


\begin{table}
  \centering
  \subfloat[8,192 OpenEDS images used in training]{%
    \begin{tabular}{c||cccc}
        \textbf{\diagbox{Model}{Dataset}} & \textbf{4,096 RITEyes} & \textbf{4,096 CGAN} & \textbf{4,096 SRCGAN (ours)} & \textbf{2,915 SRCGAN-S (ours)} \\ \hline \hline
        \textbf{RITnet} & 0.94$\pm$0.00 & \cellcolor{LightCyan}\textbf{0.94$\pm$0.00} & 0.94$\pm$0.00 & \cellcolor{LightCyan}\textbf{0.94$\pm$0.00} \\ 
        \textbf{DANN (ours)} & 0.94$\pm$0.00 & 0.93$\pm$0.00 & 0.94$\pm$0.00 & 0.93$\pm$0.01 \\ 
    \end{tabular}
  }
  \vspace{.4cm}
  \subfloat[64 OpenEDS images used in training]{%
    \begin{tabular}{c||cccc}
        \textbf{\diagbox{Model}{Dataset}} & \textbf{4,096 RITEyes} & \textbf{4,096 CGAN} & \textbf{4,096 SRCGAN (ours)} & \textbf{2,915 SRCGAN-S (ours)} \\ \hline \hline
        \textbf{RITnet} & 0.83$\pm$0.04 & 0.80$\pm$0.04 & 0.81$\pm$0.01 & 0.82$\pm$0.01 \\ 
        \textbf{DANN (ours)} & \cellcolor{LightCyan}\textbf{0.90$\pm$0.02} & \cellcolor{LightCyan}\textbf{0.89$\pm$0.01} & \cellcolor{LightCyan}\textbf{0.90$\pm$0.01} & \cellcolor{LightCyan}\textbf{0.89$\pm$0.02} \\ 
    \end{tabular}
  }
  \vspace{.2cm}
  \caption{Performance (mIoU) comparison of RITnet and DANN when training on $4{\small,}096$/$2{\small,}915$ source domain images together with $8{\small,}192$ (a), or $64$ (b) images from the OpenEDS dataset. Best performance highlighted in bold text cyan cell.} \label{fig:ritnet-dann-compare2}
\end{table}

We compare the performance of the RITnet and DANN segmentation networks when trained on datasets consisting of $4{\small,}096$ or $2{\small,}915$ synthetic eye images and a varying number of real images. Our experimental results are shown in Table~\ref{fig:ritnet-dann-compare2} for $8{\small,}192$ and $64$ real images (top and bottom). We observe that the performance of the two models is close to each other when the number of real training images is high, and the DANN module outperforms RITnet when training with a smaller number of real images (see Fig.~\ref{fig:ritnet-dann-compare} and Table~\ref{fig:ritnet-dann-compare2}). 
Therefore, we conclude that the DANN module improves the generalization between different domains, crucially closing the domain gap while increasing performance, with respect to mIoU, of the segmentation networks when fewer real images are used in the training process.




\textbf{Privacy and Ethics.} No new human data was recorded by the authors for this study. We instead utilized an existing dataset of real human eyes (OpenEDS) as our target domain. The source datasets were either rendered using computer graphics (RITEyes) or generated by neural networks (CGAN, SRCGAN, SRCGAN-S). Importantly, our work makes substantial contributions toward the objective of minimizing reliance on actual human training data. Concretely, given that we develop a modular neural system that is trained mostly on synthetic data in order to estimate the segmentation for real eyes, we reduce the need to collect eye tracking data on actual human subjects. Each method in our work also contributes to the mitigation of privacy issues/concerns. For example, while our Structure Retaining CycleGAN method needs a real eye dataset to which to map a synthetic dataset, only a small fixed number of real images are ultimately required to establish the centroid and spread of the target distribution (as compared to performant systems that require a vast collections of human eye images containing sensitive eye-tracking data). 
This reduces the risk of human data exposure and violation of the subjects' data privacy. Furthermore, avoiding/reducing the need to record human data further protects humans from exposing aspects of their identity such as facial biometric information, facial behaviors, gaze behavior, and subject personality. A positive consequence is that data sovereignty is ensured.

In Fig.~\ref{fig:ritnet-dann-compare} and Table~\ref{fig:ritnet-dann-compare2}, 
we observe that although the performance of both the RITnet and DANN segmentation networks is low when there are no real training images and performance for both  increases proportionally with respect to the number real training images, the performance of the DANN model is higher and more stable than that of RITnet once we use a small number of real training images. This further reinforces the fact that DANN exhibits the ability to generalize across domains given only a fixed, finite set of real training image samples while RITnet cannot. This, again, circumvents the need to record large amounts of training data from the actual human subjects.

While using a small, finite number of real human eye image samples as the target domain is beneficial for the reasons listed above, we acknowledge that this approach can potentially introduce issues of bias and fairness, particularly when different ages, genders, races, eye-textures, skin color, etc. are not represented in the target real dataset.

\section{Conclusion}
\label{sec:conclusion}




In this paper, we presented a multi-step neural pipeline for tackling the problem of \textit{sim2real} in the context of eye-tracking through the use of domain adaptation. Our architecture consists of three main components. First, a novel Structure Retaining CycleGAN is implemented to reconstruct synthetic eye images while ensuring they match the distribution of real eye images. Second, a Siamese Network is designed to filter out poorly-reconstructed images through a learned dataset pruning approach. Lastly, a model adapted from a domain-adversarial neural network structure is employed to semantically segment the real images.

Subjectively, the datasets reconstructed at the different stages as we progress through our pipeline do appear to be more realistic/plausible (see Fig.~\ref{fig:sample-dataset}). Our objective results further indicate that the later stage datasets do indeed yield greater performance (see Fig.~\ref{fig:srcgan-miou} and Table~\ref{fig:srcgan-miou2}). Specifically, the SRCGAN dataset outperforms CGAN in terms of mean distance to the real centroid and downstream mIoU score. The SRCGAN-S dataset performance is similar to SRCGAN but offers the additional benefit of faster training time since the number of images in the dataset is greatly reduced compared to SRCGAN ($2{\small,}915$ for SRCGAN-S vs. $9{\small,}216$ for SRCGAN, see Table~\ref{tab:num-images}). The fact that we are able to achieve similar performance with fewer images confirms that our Siamese Network successfully filters out problematic synthetic image samples. Finally, our results show that our proposed DANN segmentation network outperforms RITnet in terms of segmentation mIoU score when only a small number of real images is included among the synthetic datasets used for training (see Fig. \ref{fig:ritnet-dann-compare} and Table~\ref{fig:ritnet-dann-compare2}). Overall, we have provided empirical evidence that our multi-step neural architecture results in improved synthetic datasets for training semantic segmentation models, and we also present an improved segmentation model (DANN). Furthermore, our results have positive implications for reducing the cost and burden associated with capturing and manually labeling large quantities of real human eye data, which in turn also promotes data privacy. 

In terms of future research work, the results we presented focus on the overall mIoU score averaged across distinct eye regions. However, delving into region-specific mIoU scores could yield additional insights and improvements. For instance, our observations indicate that the mIoU score for the sclera region tends to lag behind those for the pupil and iris regions. Further exploration of region-specific objective functions may effectively address and enhance performance in these specific areas. Additionally, in our DANN sub-module, the reverse gradient from the domain classifier currently flows through the encoder of the segmentation network only (see Section~\ref{subsec:dann-eye}). This makes the domain generalization learning occur in the encoder but not in the decoder (the ``segmenter''). It is worth exploring whether this might be a contributing factor in the DANN model's slight underperformance compared to RITnet when the number of real images gets (much) larger. 
Another challenge central to the problem of domain adaptation relates to the ability to generalize eye segmentation models with respect to the target domain. Although the real target domains such as the OpenEDS dataset consider different combinations of age ($19$-$65$), sex (male/female), usage of glasses, and corneal topography~\cite{openeds}, they may not account for the diversity in other features such as sclera/iris texture. While it is possible to introduce diversity in the rendering pipeline by using different head models representing different genders, ages, races, iris/sclera textures, etc. as in RITEyes~\cite{riteyes}, our current pipeline does not prioritize the retention of these features during domain transfer. Therefore, it is worthwhile to explore the development of domain transfer functions that preserve such features.
Additionally, while mapping from the synthetic to the real domain, it may be beneficial to incorporate eye glint as an additional transferable structure in order to further improve the realism of the resulting eye images. While our current research integrates glints within the eye region textures, future studies could explore framing it as a separate structural element and aim to preserve its characteristics from the source domain. Lastly, future works can also explore the impacts of different image resolution on the efficiency of our algorithms.



\bibliographystyle{acm}
\bibliography{main}
\end{document}